\documentclass[graybox, a4paper]{svmult}

\usepackage{type1cm}        
\usepackage{makeidx}         
\usepackage{graphicx}        
\usepackage{multicol}        
\usepackage[bottom]{footmisc}

\usepackage{newtxtext}       %
\usepackage{newtxmath}       

\usepackage[ruled,vlined, linesnumbered]{algorithm2e} 
\SetKwInOut{Parameter}{Parameters}

\usepackage{float}
\usepackage{enumitem}

\usepackage{subcaption}
\usepackage{caption}

\makeindex             

\usepackage{xcolor}
\newcommand{\acro}{RMF} 

\newcommand{\kyon}[1]{{\color{cyan} Kyon: #1}}

\newcommand{\pr}[1]{}  
\newcommand{\ph}[1]{} 

\begin{document}

\title*{Autonomous Off-road Navigation over Extreme Terrains with Perceptually-challenging Conditions}
\titlerunning{Autonomous Off-road Navigation}

\author{Rohan Thakker$^{1*}$, Nikhilesh Alatur$^{2*}$, David D. Fan$^{1*}$, Jesus Tordesillas$^{3*}$, Michael Paton$^{1}$, Kyohei Otsu$^{1}$, Olivier Toupet$^{1}$, and Ali-akbar Agha-mohammadi$^{1}$}

\institute{$^{*}$The authors contributed equally.\\$^{1}$Jet Propulsion Laboratory (JPL), California Institute of Technology (Caltech), 4800 Oak Grove Dr. Pasadena, CA, \email{\texttt{firstname.lastname@jpl.nasa.gov}}\\
$^{2,3}$ Work done while at the JPL, Caltech. Currently at $^{2}$Swiss Federal Institute of Technology, Zurich \email{\texttt{alaturn@ethz.ch}}, $^{3}$Massachusetts Institute of Technology \email{\texttt{jtorde@mit.edu}}\\
       {\small \textcopyright 2020, California Institute of Technology. All Rights Reserved}}

\authorrunning{Thakker et al.}
%
%
\maketitle

\vskip -2.2cm

\abstract{
We propose a framework for resilient autonomous navigation in perceptually challenging unknown environments with 
mobility-stressing elements such as uneven surfaces with rocks and boulders, steep slopes, negative obstacles like cliffs and holes, and narrow passages. 
Environments are GPS-denied and perceptually-degraded with variable lighting from dark to lit and obscurants (dust, fog, smoke).
Lack of prior maps and degraded communication eliminates the possibility of prior or off-board computation or operator intervention. 
This necessitates real-time on-board computation using noisy sensor data.
To address these challenges, we propose a resilient architecture that exploits redundancy and heterogeneity in sensing modalities.
Further resilience is achieved by triggering recovery behaviors upon failure.
We propose a fast settling algorithm to generate robust multi-fidelity traversability estimates in real-time.
The proposed approach was deployed on multiple physical systems including skid-steer and tracked robots, a high-speed RC car and legged robots, as a part of Team CoSTAR's effort to the DARPA Subterranean Challenge, where the team won 2\textsuperscript{nd} and 1\textsuperscript{st} place in the Tunnel and Urban Circuits, respectively.
}
\begin{figure}[h!]
\centering
\begin{subfigure}{0.9\textwidth}
  \centering
  \includegraphics[width=0.45\linewidth]{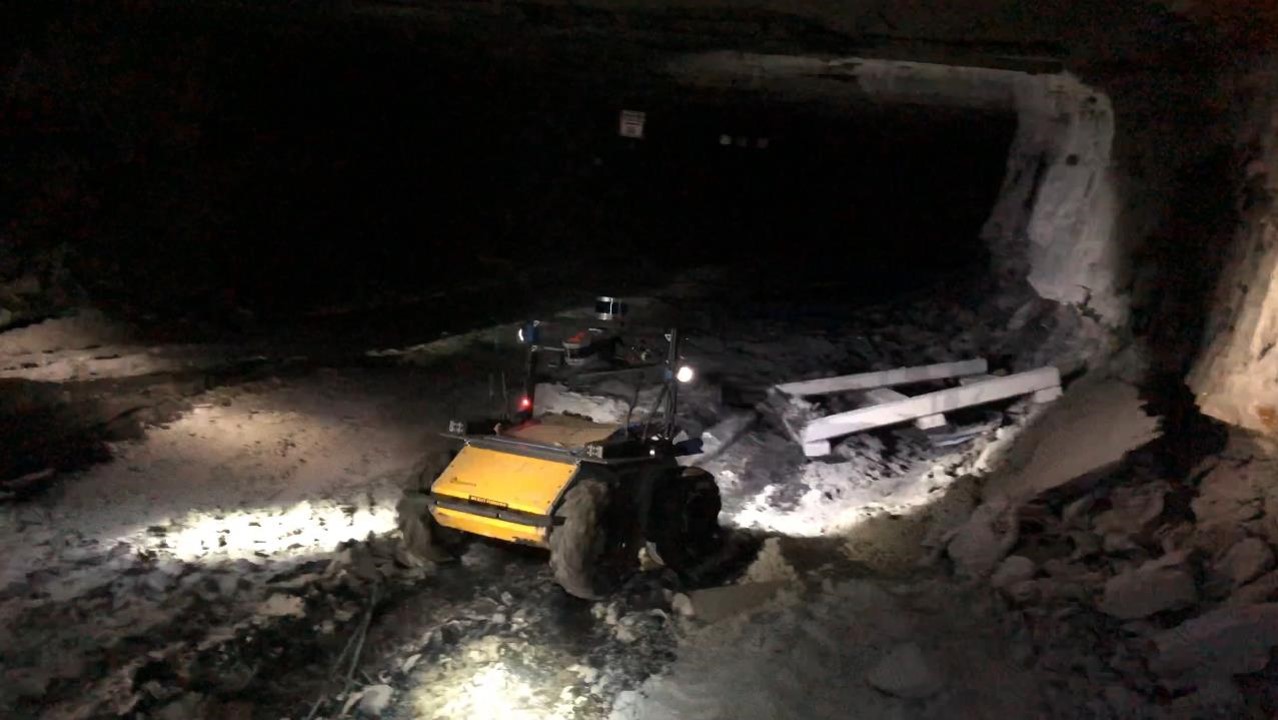}
  \includegraphics[width=0.45\linewidth, trim={0 29mm 0 0}, clip]{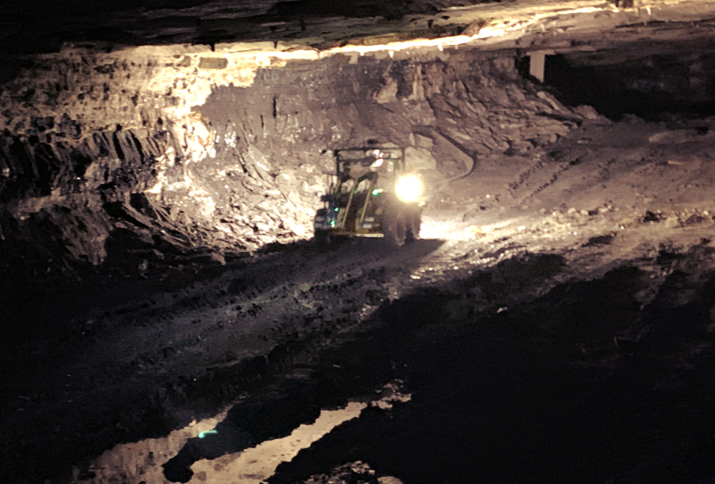}  
  \caption{Rough Terrain}
  \label{fig:sfig1}
\end{subfigure}%
\hspace{0.04cm}
\begin{subfigure}{.3\linewidth}
  \centering
  \includegraphics[width=1.0\linewidth]{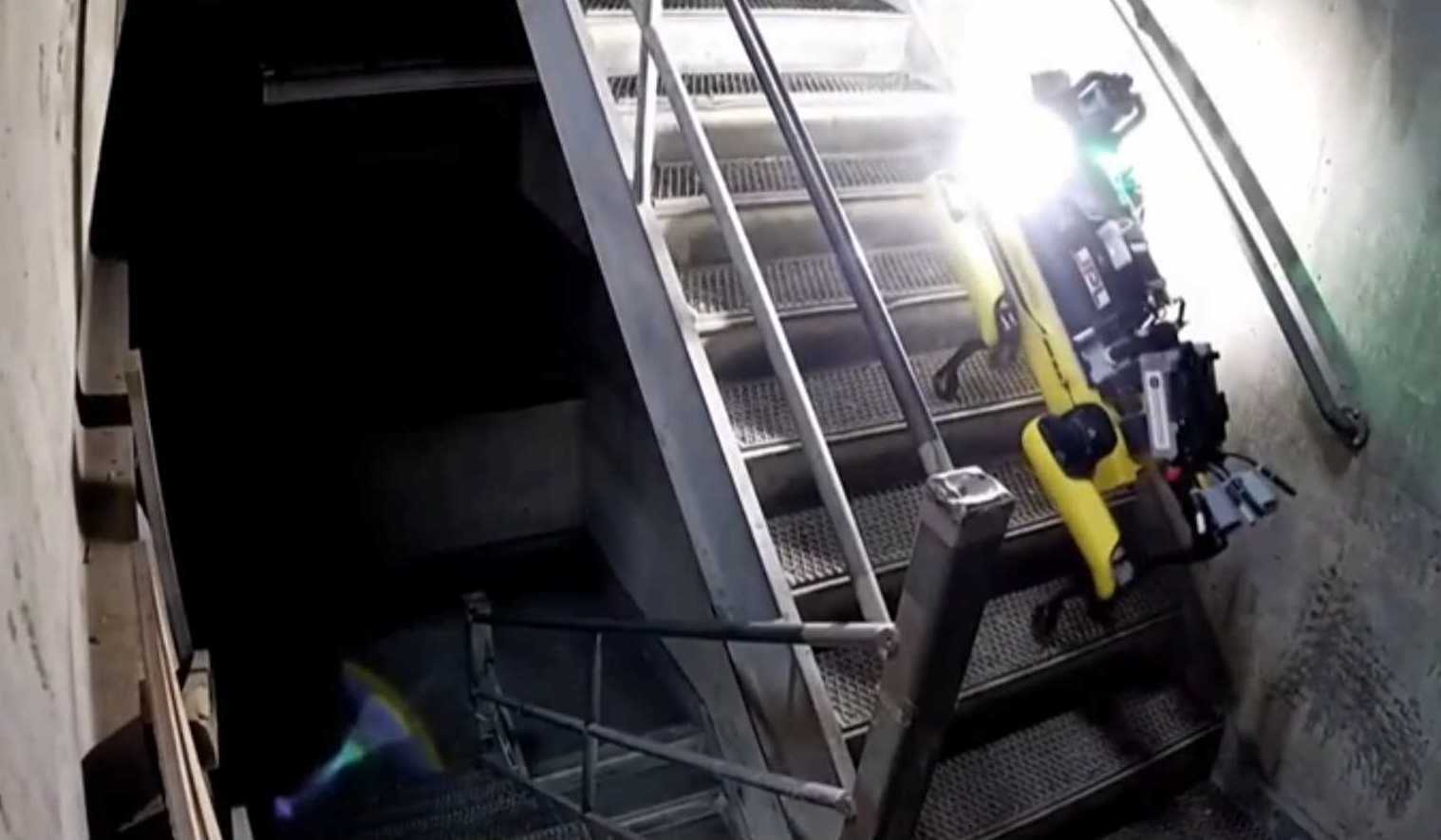}
  \caption{Stairs (Legged)}
  \label{fig:sfig2b}
\end{subfigure}
\begin{subfigure}{.3\textwidth}
  \centering
  \includegraphics[width=\linewidth, trim={0 1.5cm 0 1cm}, clip]{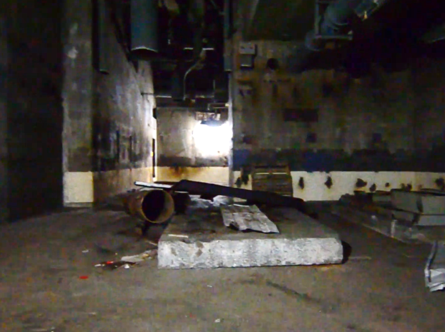}
  \caption{Cluttered environment}
  \label{fig:sfig4}
\end{subfigure}%
\hspace{0.01cm}
\begin{subfigure}{.3\linewidth}
  \centering
  \includegraphics[width=\linewidth, trim={0 0 0 2.5cm}, clip]{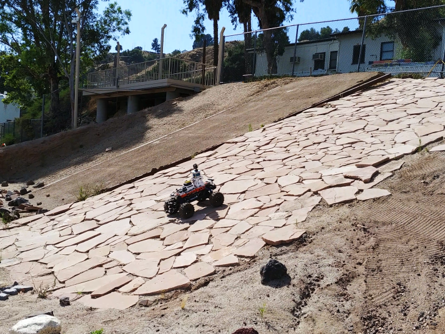}
  \caption{Sloped terrain}
  \label{fig:sfig2a}
\end{subfigure}
\vspace{-5mm}
\end{figure}
\begin{figure}[ht]\ContinuedFloat
\begin{subfigure}{.32\textwidth}
  \centering
  \includegraphics[width=1.0\linewidth, trim={0 0 0 0.4cm}, clip]{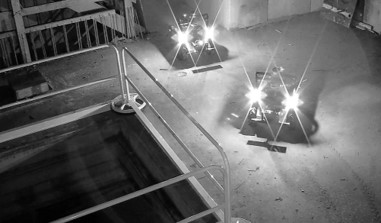}
  \caption{Negative Obstacles}
  \label{fig:sfig5b}
\end{subfigure}
\begin{subfigure}{.32\textwidth}
  \centering
  \includegraphics[width=1.0\linewidth]{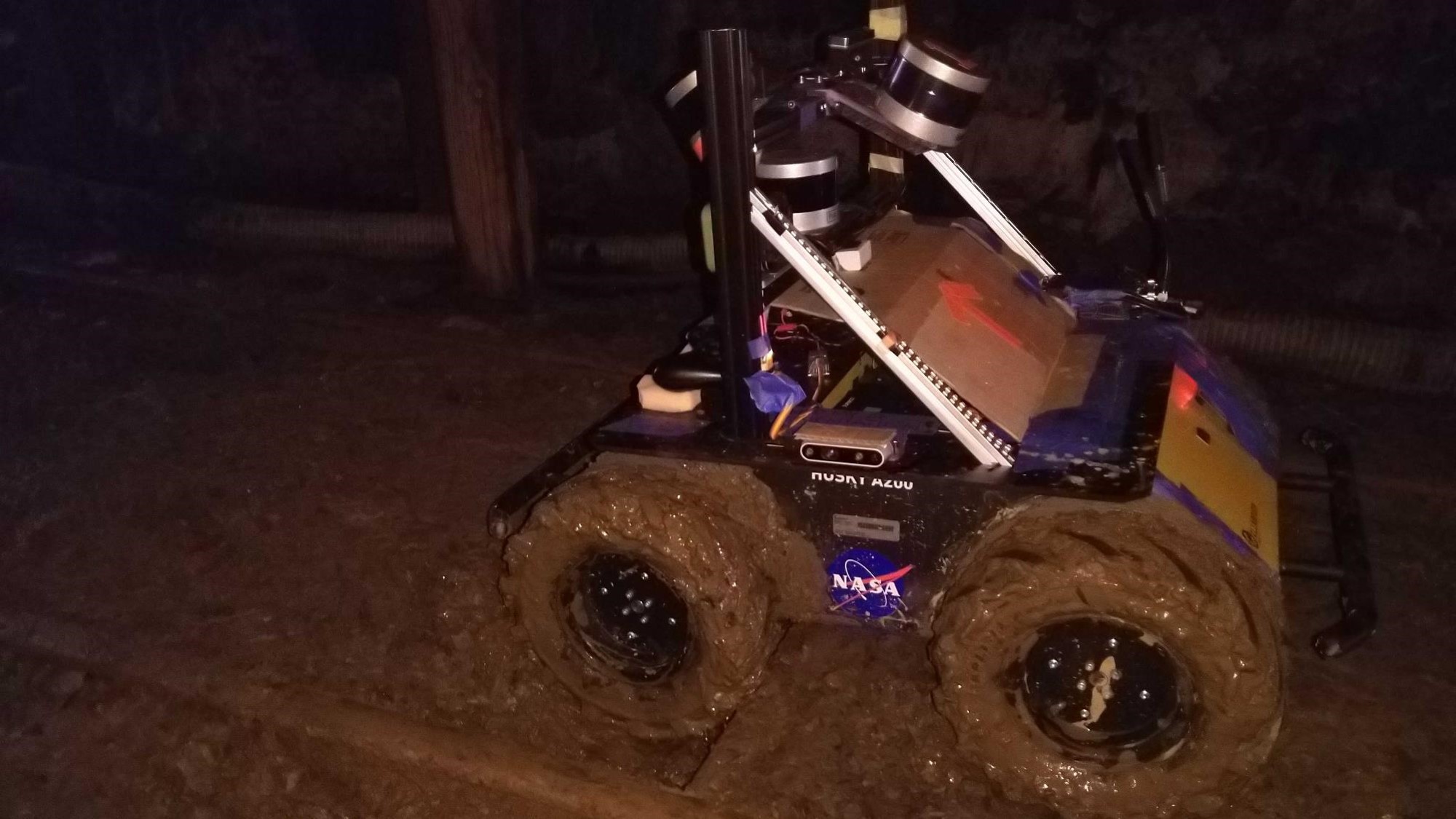}
  \caption{Mud}
  \label{fig:sfig3}
\end{subfigure}%
\hspace{0.1mm}
\begin{subfigure}{.32\textwidth}
  \centering
  \includegraphics[width=\linewidth, trim={0 0.7cm 0 0}, clip]{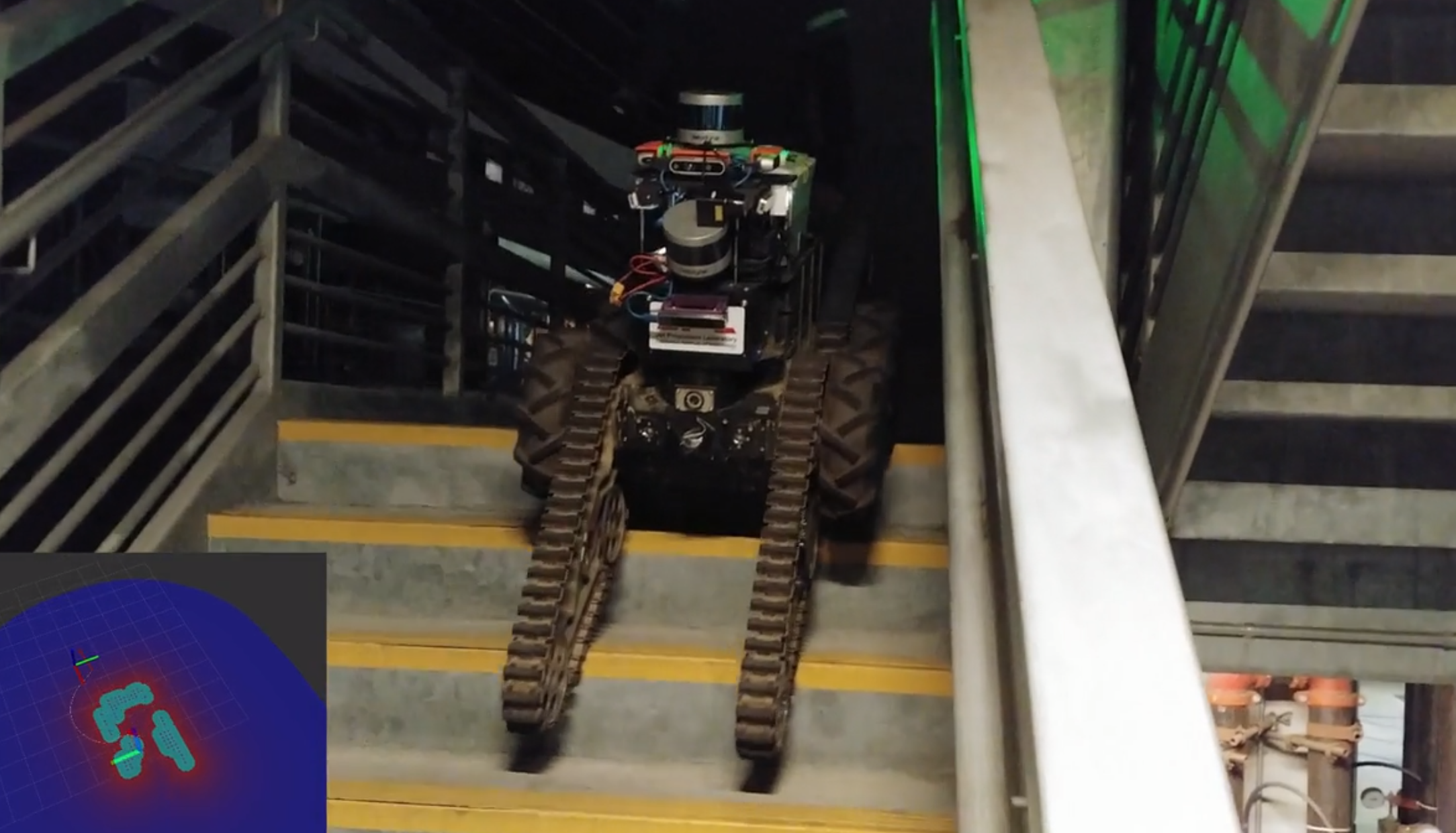}
  \caption{Stairs (Tracked)}
  \label{fig:sfig5d}
\end{subfigure}
\caption{Mobility-stressing components at test sites: a) Arch Coal Mine, WV, b,c,e) Powerplant in Satsop Business Park, WA,  d) Mars Yard, NASA-JPL, CA,and f) Beckley Exhibition Mine, WV. 
Robots demonstrated long-range autonomous navigation capability, such as in Arch Mine (1100m), Beckley Exhibition Mine (1000m).
}
\label{fig:traversability_challenges}
\vspace{-7mm}
\end{figure}
\textbf{Supplementary material}: A video is available at \textcolor{blue}{\url{https://youtu.be/cjxN2Tu-qVY}}
\vspace{-7mm}
\section{Introduction}

\label{sec:intro}
Autonomous off-road navigation over extreme terrains and perceptually-degraded environments has a wide range of applications from space exploration \cite{haruyama2012lunar} to terrestrial search and rescue, including subterranean (tunnels, mines, caves) exploration \cite{murphy2014disaster}. 
This poses a set of \textbf{key challenges}:

\begin{enumerate}[label={[C\arabic*]},align=left]
    \item \label{c:start} \label{c:extereme_terrain} 
    \textit{Negotiating mobility-stressing elements} such as rough, high-sloped, and cluttered terrains with stairs and narrow passages while avoiding hazards such as rocks, overhangs, and negative obstacles including pits, cliffs, etc. (Fig.~\ref{fig:traversability_challenges}).
    
    \item \label{c:perception}\textit{Perceptually degraded GPS-denied environments} require the robots to cope with large uncertainties and degraded sensing due to presence of obscurants (dust, fog, smoke), low-light conditions, lack of features/landmarks, motion blur and issues with dynamic range (Fig.~\ref{fig:costmap_arrows2}).
    \item \label{c:autonomus}\textit{Resilient self-reliant operations in comms-denied environment} without any operator intervention are required due to degraded communication.
    The robot is also required to return back to communication range if the mission is infeasible or violates the desired risk posture.
    \item \label{c:realtime}\textit{On-board operations in unknown environments} at high-speeds are required since lack of prior maps and degraded communications eliminate any possibility of prior or off-board processing and finite mission time constraints make operations at maximum speed supported by hardware, desired. 
    \item \label{c:end}\label{c:scalability}\textit{Scalability to different environments and robots}: such as tracked, wheeled and legged which have disparate speed and traversability abilities (Table.~\ref{table:config_exp_platforms}).
\end{enumerate}

\textbf{Related work:}
Papadakis \cite{papadakis2013trav} summarises the prior work on traversability estimation by classifying them in proprioceptive methods and exteroceptive methods.

\cite{Otsu2019} performs fast, approximate settling by computing upper and lower bounds for the settled pose.
In \cite{krusi2017driving}, the authors use principal component analysis on the local surface to estimate the settled pose.
The settling algorithms can operate on instantaneous scans \cite{InstPcWafr2016} which are robust to localization noise or by fusing multiple scans \cite{nanomap2018} that provide a longer range.

The work in \cite{Brunner2015Trav} runs an iterative algorithm to estimate the contact points between an articulated, tracked robot and the terrain mesh \cite{BrunnerAstar} for settling.
\cite{Ma2019} proposes a computationally fast method to run a physics simulation that drops the robot under the influence of gravity onto the surface, which is represented as B-Patch \cite{Iagnemma2008}. 

\cite{fankhauser2018probabilistic} provides a probabilistic formulation 
by extending a discrete elevation map \cite{Goldberg2002} 
to account for uncertainty in localization.
 
The objective of this work is to develop a holistic solution that can address all challenges \ref{c:start}-\ref{c:end}, simultaneously.

\textbf{Contributions:}
\begin{enumerate}
    \item  To address these challenges, we propose a Resilient Multi-fidelity Architecture (\acro) design which shows how these different prior works can be combined to form a resilient end-to-end system.
    \acro \ leverages a multi-fidelity traversability estimator which generates accurate estimates in short-range by using instantaneous scans and approximate estimates in mid-range and long-range by using a temporal map. 
    Furthermore, it is designed to eliminate single-point failures and achieve further resiliency by exploiting redundancy and heterogeneity of different sensing modalities \ref{c:perception}. 
    \item We propose a fast settling algorithm that estimates traversability in extreme terrains \ref{c:extereme_terrain} for multiple robot types \ref{c:scalability} in real time \ref{c:realtime}.
    \item We propose a sequence of field-tested autonomous recovery behaviors to eliminate human interventions \ref{c:autonomus}.
    \item We push the boundaries of state-of-practice and demonstrate the proposed solution through deployment on a variety of ground vehicles including wheeled (Ackermann and skid-steered), tracked, and legged systems, by conducting intense field testing in various subterranean environments (Fig. \ref{fig:traversability_challenges}, Table \ref{table:config_exp_platforms}).
\end{enumerate}

The proposed architecture was successfully deployed by Team CoSTAR\footnote{https://costar.jpl.nasa.gov/} as our local navigation solution for large-scale, subterranean exploration and mapping missions during the DARPA Subterranean (SubT) Challenge\footnote{https://www.subtchallenge.com/}. 
Together with other autonomy modules for large-scale localization and mapping \cite{Kamak2020icra}, resilient state estimation \cite{hero2019isrr} and perception-aware global planning \cite{Agha-mohammadi2014}, it formed the overall autonomy solution called \emph{NeBula} (Networked
Belief-aware Perceptual Autonomy) \cite{Amanda2020iros} that led to Team CoSTAR winning 2\textsuperscript{nd} place in the Tunnel Circuit and 1\textsuperscript{st} place in the Urban Circuit of the SubT Challenge.

\section{Methodology}
\label{sec:methodology}

We push the boundaries of the state-of-practice by enabling on-board navigation in complex rough terrain under severe perceptual, computational and real-time constraints. In this section, we discuss an architecture and supporting techniques of the proposed system.

\vspace{-5mm}

\subsection{Resilient Multi-fidelity Architecture}
\vspace{-10mm}
\begin{figure}[ht]
\centering
\includegraphics[width=\textwidth]{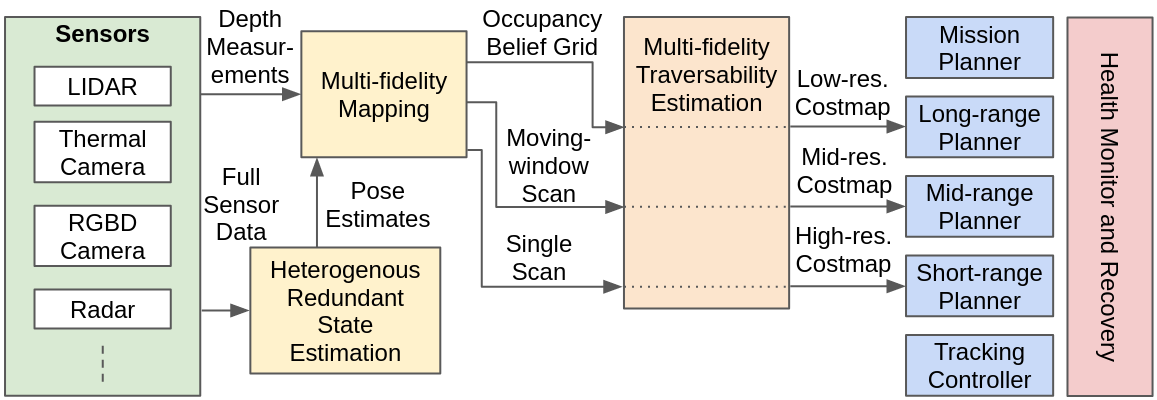}
\caption{Resilient Multi-Fidelity (RMF) Architecture.}
\label{fig:signal_flow_diagram}       
\vspace{-4mm}
\end{figure}

\textbf{Heterogeneity and Redundancy of Sensors to Eliminate Single-Point Failures}:
Fig. \ref{fig:signal_flow_diagram} shows our pipeline to generate the multi-fidelity traversability cost estimates for motion planning. 
To design a resilient system capable of extreme terrain navigation, we start by building a stack of sensors with heterogeneous modalities such as lidar, radar, thermal and visible cameras, etc. 
Further redundancy is achieved by adding multiple sensors of the same modality, pointing in different directions to achieve larger coverage and eliminate single point failures.

\textbf{Heterogeneous Resilient State Estimation}:
These sensors signals are process by our estimation module HeRO (Heterogeneous Redundant Odometry) \cite{hero2019isrr} that further adds additional analytical redundancy by running multiple odometry algorithms \cite{ebadi2020lamp}, \cite{kramer2020radar}, \cite{lion2020}, performs confidence checks to quantify uncertainty and finally produces robust pose estimates by multiplexing to the most robust odometry source.
This provides resiliency to state estimation failures which are usually triggered by many factors like dust, low-light conditions, lack of features/landmarks, motion blur and issues with dynamic range.

\textbf{Robust Multi-Fidelity Traversability Estimation in Real-time}:
To achieve further robustness to noise in perception, we perform traversability estimation at three different levels.
First, a high-resolution short-range traversability costmap is constructed using the most recent depth scan which does not suffer from localization noise. 
This allows robust high-fidelity traversability estimation by preventing the propagation of localization noise into the planning and control layers.
Second, mid-resolution mid-range planner uses a moving window of last $N$ depth scans for traversability estimation which allows coverage over a larger region.

While these depth estimates are sensitive to localization noise, the traversability estimation is less susceptible to this noise since it is performed at a lower fidelity than the high-resolution costmap. 
Specifically, features which are within the magnitude of perception noise are ignored (e.g. rocks smaller than 30 cm).
Third, the low-resolution long-range costmap uses an occupancy belief grid that maintains a probability distribution of occupancy which is a sufficient statistic of all past depth measurements \cite{agha2017CRM}-\cite{hornung2013octomap}.
This allows for incorporating highly noisy yet informative measurements at long ranges.  The low-resolution costmap achieves robustness to this noise by estimating traversability at the least fidelity (e.g. only detecting large obstacles such as walls).
This multi-fidelity approach is an efficient and practical way of handling perception noise without adding too much computational burden.
In the next subsection, we describe the details of the settling algorithm used to estimate traversability.

\textbf{Real-time On-board Planning and Control:}
To achieve real-time performance with limited on-board computation, the planning and controls problems is decoupled into several hierarchical layers. 
The mission planner keeps track of high-level states such as battery status, time remaining, coverage goals, etc. \cite{otsu2020supervised} and generates a desired goal for the long-range planner.
The long-range planner finds a path on a long-range roadmap (0-10\,km range) and generates a goal for the mid-range planner.
The mid-range planner performs search on the mid-resolution costmap (0-50\,m range) to obtain a geometric path to this goal which is tracked by the local kinodynamic planner by generating an optimal trajectory from a set of motion primitives prioritized by using the high-resolution costmap (0-10\,m range).
Finally, the low-level tracking controller tracks this kinodynamic trajectory.
\cite{Amanda2020iros} describes the details of how this was implemented for a legged robot.

\textbf{Resilience at System-level:} Note that while the long-range and mid-range planner can produce intermittent unsafe goals due to presence of localization errors, our system is still resilient to these failures since the short-range planner closes the loop using only instantaneous or near-instantaneous scans (past <1 second), thus eliminating past errors in localization.
The health monitor asynchronously monitors the system state to trigger recovery behaviors when failures are detected.

\subsection{Settling-Based Traversability Analysis} \label{subsec:Settling-Based}

\begin{figure}[t]   
\includegraphics[width=\textwidth]{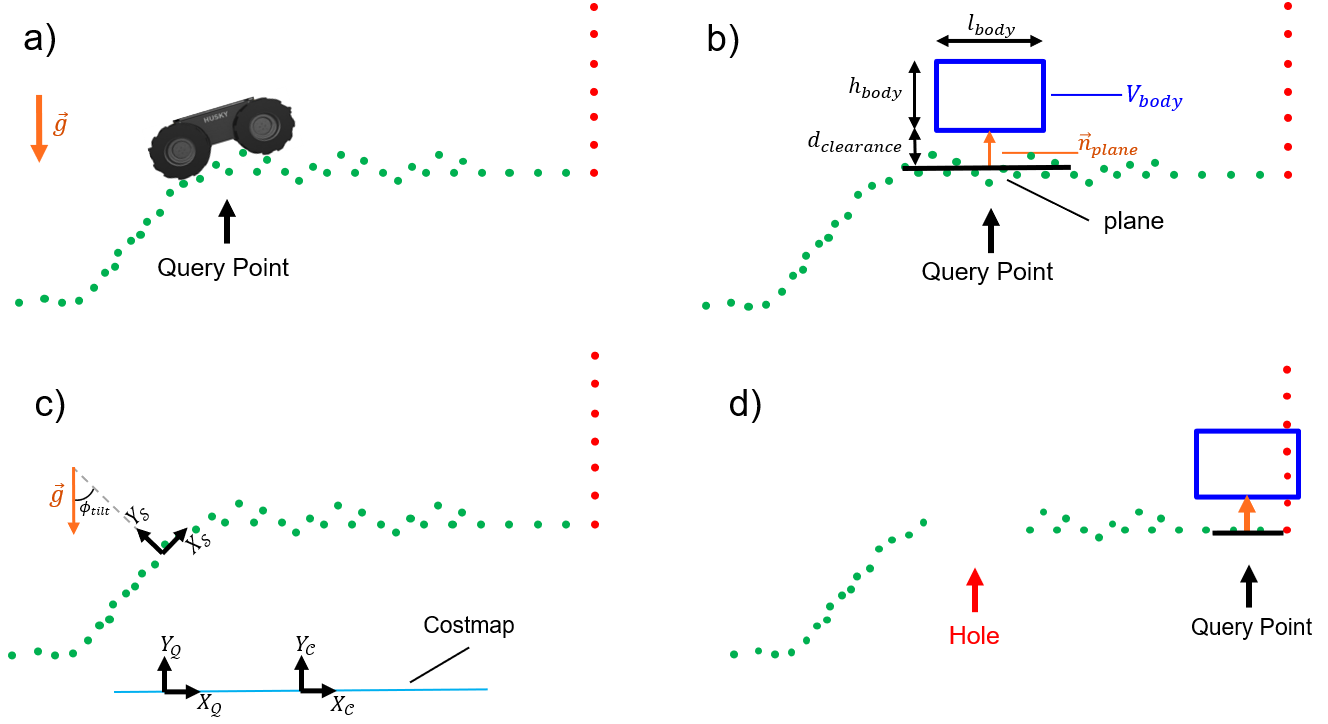}
\caption{Settling-based traversability: a) settling, b) body collision model, c) stability check, d) collision detection and negative obstacle detection using low point density.}
\label{fig:trav_concept}       
\vspace{-4mm}
\end{figure}

\ph{Algorithm Overview}
This section describes how traversability is assessed over a point cloud for a single query pose.
The traversability assessment for a query pose consists of three steps.
First, the input surface point cloud is segmented into ground and obstacle points. Second, the robot is virtually \emph{settled} on a ground point cloud at a given query pose.
Third, various traversability metrics are computed from the settled pose, surface point cloud, and its interaction.

\ph{Point cloud segmentation} The algorithm begins by first segmenting the surface point cloud into ground and obstacle points. In rough terrain (which is a target of this paper), the boundary between ground and obstacles is vague. To have a consistent method across different environments, we set a following definition based on the mechanical capability of the robot platform: ground points are surface points of an area where the robot can mechanically drive, and obstacle points are remainder points that locate above the ground points. Multiple implementations are possible for this segmentation task. We employed a simple geometric approach based on line-fitting \cite{himmelsbach2010fast}.

\ph{Settling}
After extracting the ground point cloud, a robot model is placed on arbitrary locations on the point cloud to evaluate traversability. The process is depicted in Fig.~\ref{fig:trav_concept}. A settling algorithm is used to solve for an $SE(3)$ pose of the robot given an $SE(2)$ query pose. Different settling methods should be considered based on the platform types. One approximation method that is applicable to many ground vehicles is to fit a plane using points under the robot footprint. The normal of fitted plane is used to approximately estimate the orientation of settled pose and derive the placement of the body collision model.

\ph{Traversability metrics}
Once a full pose is determined for a query pose via settling, the following metrics are computed using the pose and point cloud. The final traversability is evaluated based on the combination of these metrics:
\begin{itemize}

    \item \textit{Tip-over stability}: The robot must avoid steep slopes, where it could tip over.
In theory, the robot's footprint and orientation highly influence the stability of the robot.
In \cite{Brunner2015Trav}, the authors list several stability metrics that can be used for stability checking.
However, for robots whose footprint have an aspect ratio close to one, one can simplify the stability analysis by applying a threshold on the angle between the surface normal and the gravity vector. This stability check is depicted in Fig.~\ref{fig:trav_concept}c.

    \item \textit{Positive obstacles}: Positive obstacles, such as rocks below the robot's belly, overhangs, walls, or larger obstacles, may collide with robot's body. The surface points that interfere with the robot's collision model are counted, and if the number of colliding points is larger than a threshold, that location is marked as untraversable, see Fig.~\ref{fig:trav_concept}d.

    \item \textit{Negative obstacles}: Negative obstacles include cliffs and holes, which can severely damage the vehicle if they are not avoided. We create two categories based on sensor coverage. If the shape of negative obstacle is fully visible, the previous slope and collision checks are applied to evaluate if it is safe to drive. If the negative space is occluded or not visible, we check the density of points around the negative space.  If too few ground points are available, we assume the worst case and mark the area as untraversable.
    
\end{itemize}

By changing the size of the robot bounding box used in the settling algorithm, we obtain different fidelities of traversability estimates.  A larger bounding box will generate higher fidelity estimates which can detect hazards such as very small rocks, while a smaller bounding box will result in lower fidelity estimates to detect large hazards such as walls.
A larger size will be more sensitive to localization noise and hence is only applied to near-instantaneous measurements, whereas a smaller size is robust to localization noise and can be applied to temporally fused depth maps.

\subsection{Recovery Behaviors}

In the real world, failures are unavoidable. Some failures (e.g., mechanical failure) are fatal, while others (e.g., algorithmic failure) can potentially be resolved autonomously, allowing the robot to continue the mission. Hence, as a last line-of-defense, we design behaviors to recover the system from non-fatal failures. The following recovery behaviors are implemented and triggered when relevant failures are detected:
\begin{itemize}
    \item Clear the map and start building it from scratch. 
    \item Backtrack the traversed path.
    \item Move the robot in an open loop to the direction of maximum clearance.
    \item Trigger wall following (which does not require global localization).
\end{itemize}

\section{Experiments}
\label{sec:experiments}

The proposed system has been extensively tested and validated in many different natural and man-made environments, including coal/gold mines, power plants, school/office buildings, and the JPL Mars Yard; as shown in Fig.~\ref{fig:traversability_challenges}.
The navigation system was successfully used by Team CoSTAR at the DARPA SubT Challenge Circuit events (Tunnel and Urban), where the robots autonomously explored more than a kilometer in each run.

\subsection{Hardware and Software}

\begin{table}[t]
\centering
\includegraphics[width=\textwidth]{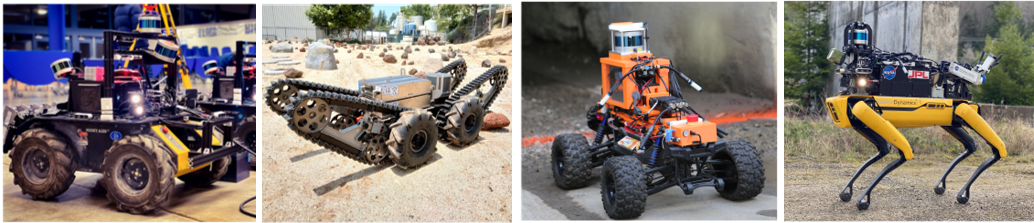}
\begin{tabular}{ | c || c | c | c | c | }
 \hline
 \textbf{Specification} & \textbf{Skid-steer$^1$} & \textbf{Tracked$^3$} & \textbf{Ackermann$^2$} & \textbf{Quadruped$^1$}\\
 \hline
 \textbf{Model} & Husky A200 & Telemax Pro & X-Maxx & Spot \\
 \textbf{Max. speed} & 1.0 m/s & 1.1 m/s & 22 m/s & 1.6 m/s \\
 \textbf{Local planner}&   DWA \cite{dwa_burgard}  & DWA \cite{dwa_burgard}& TEB \cite{Rosmann2017}&Proprietary\\
 \textbf{Controller}& PID  & PID   & PID & Proprietary\\
 \hline
 \textbf{Distance traveled} & 5.98 km & 0.76 km & 0.39 km & 2.85 km \\
 \textbf{Avg. speed during traverse} & 0.93 m/s & 0.57 m/s & 0.83 m/s & 0.65 m/s \\
 \textbf{Autonomous recoveries / km} & 6.9 & 0.16 & 0.0 & 13.0\\
 \textbf{Critical failures / km} & 0.2 & 0.0 & 0.0 & 1.1 \\
 \hline
\end{tabular}

\caption{Specification of the different robots.  Superscript on the robot type indicates location/source of data used for computing statistics: $^1$DARPA SubT Urban Competition, $^2$DAPRA SubT Tunnel Competition, $^3$Beckley Exhibition Coal Mine, WV.}
\label{table:config_exp_platforms}
\vspace{-4mm}
\end{table}

\ph{Hardware platforms and differences}
The experimental platforms are shown at the top of Table \ref{table:config_exp_platforms}.
The platforms were heavily customized with a heterogeneous sensor suite, computing units, batteries and speed controllers.
The odometry sources consist of wheel-inertial odometry (WIO), lidar-inertial odometry (LIO), visual-inertial-odometry (VIO), thermal-inertial-odometry (TIO) and kinematic-inertial-odometry (KIO). 
HeRO \cite{hero2019isrr} monitors the different inputs and selects the most reliable source to produce a continuous state estimate under perceptually degraded conditions.
\ph{Software Implementation} The proposed architecture is implemented with the ROS navigation framework\footnote{http://wiki.ros.org/navigation}.

\subsection{Results}

\begin{figure}[ht]
\centering
\includegraphics[width=\textwidth]{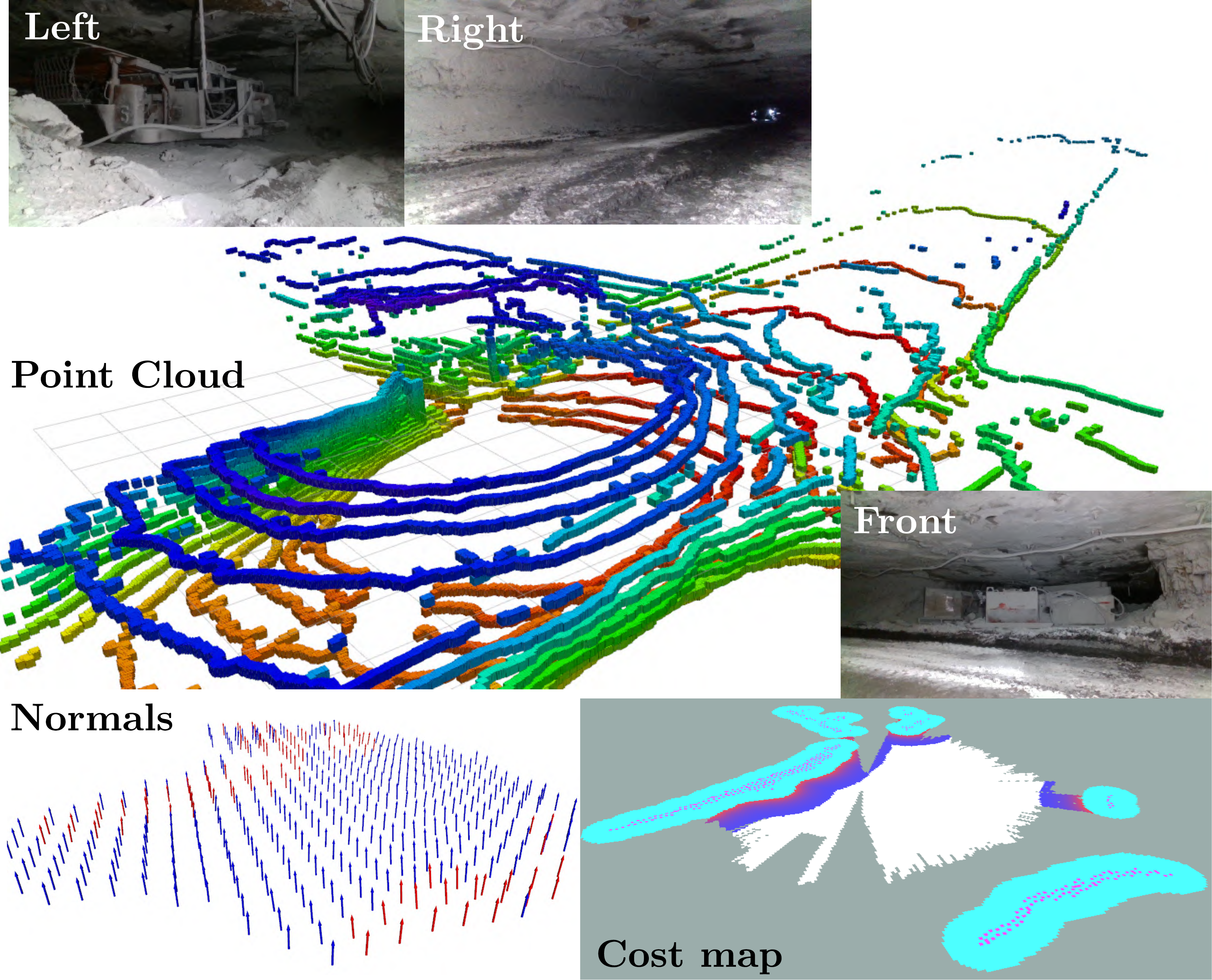}
\caption{Traversability estimation pipeline visualization showing Point Cloud from sensors, normal vector of settled robot pose (hazards marked in red) and costmap at Arch Coal Mine.}
\label{fig:costmap_arrows2}
\vspace{-4mm}
\end{figure}

\begin{figure}[ht]
\centering
\includegraphics[width=\textwidth]{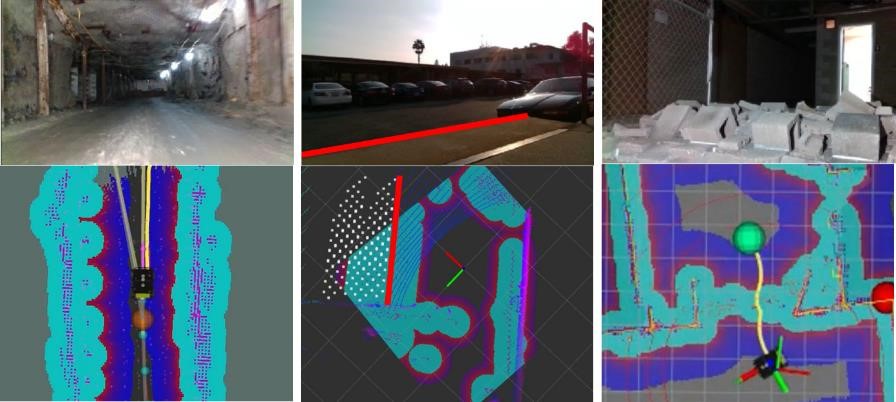}
\caption{Left to Right: Positive Obstacles, Negative Obstacles, Rubble}
\label{fig:results}       
\end{figure}

\begin{figure}[ht]
\centering
\includegraphics[width=\textwidth]{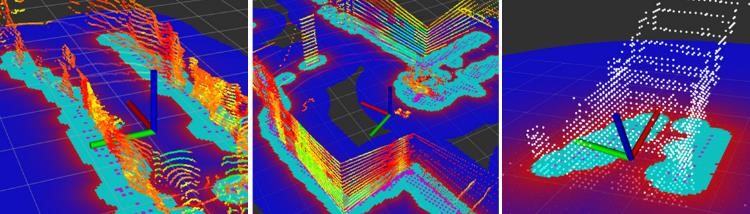}
\caption{Left to Right: Skid-Steer in Mine, Quadruped in Power Plant, Tracked on Stairs}
\label{fig:results_other}       
\vspace{-2mm}
\end{figure}

The internals of the proposed algorithm are visualized in Figure \ref{fig:costmap_arrows2}, for a mission in the Arch Coal Mine in Beckley, WV.
The costmaps for different traversability challenges, including positive obstacles, negative obstacles and rubble, are shown in Figure \ref{fig:results} for the wheeled platform A.
Figure \ref{fig:results_other} visualizes the costmap for the Ackerman platform in a gold mine, for the quadruped in the power plant and for the tracked platform on stairs. 
With a modified settling algorithm \cite{Otsu2019}, we were able to demonstrate autonomous stair climbing with the tracked platform.
The difference boils down to a more conservative stability assessment that considers the attainable worst case attitudes instead of average attitudes that result from simple plane fitting.
The addition of conservatism is motivated by the higher slope and greatly reduced contact area, which makes stairs a very challenging terrain to drive on.

\begin{figure}[t]
\centering
\begin{subfigure}{.27\linewidth}
  \centering
  \includegraphics[width=\linewidth]{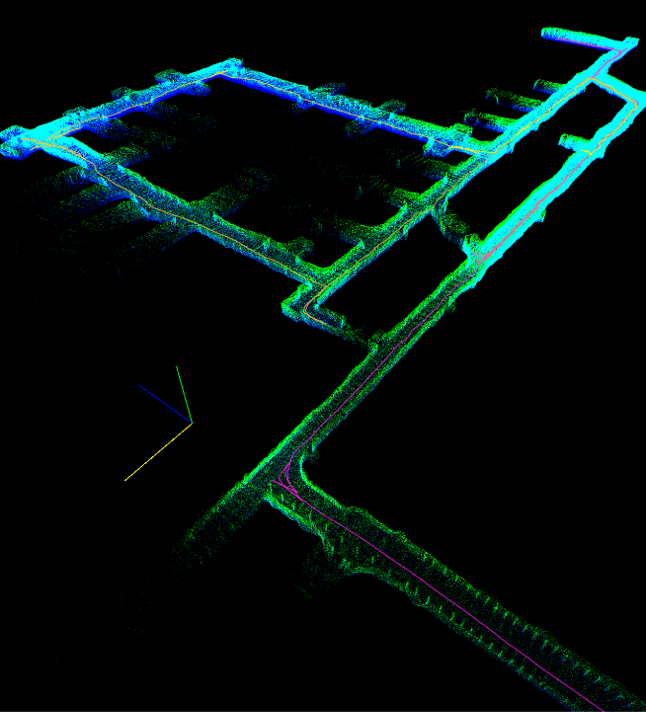}
\end{subfigure}%
\hspace{0.0cm}
\begin{subfigure}{.27\linewidth}
  \centering
  \includegraphics[width=\linewidth]{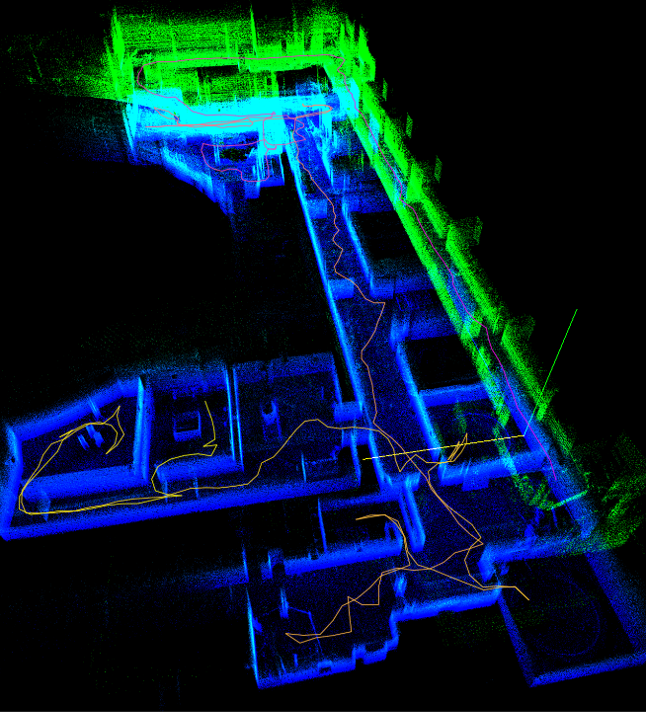}
\end{subfigure}
\hspace{-0.82mm}
\begin{subfigure}{.27\linewidth}
  \centering
  \includegraphics[width=\linewidth]{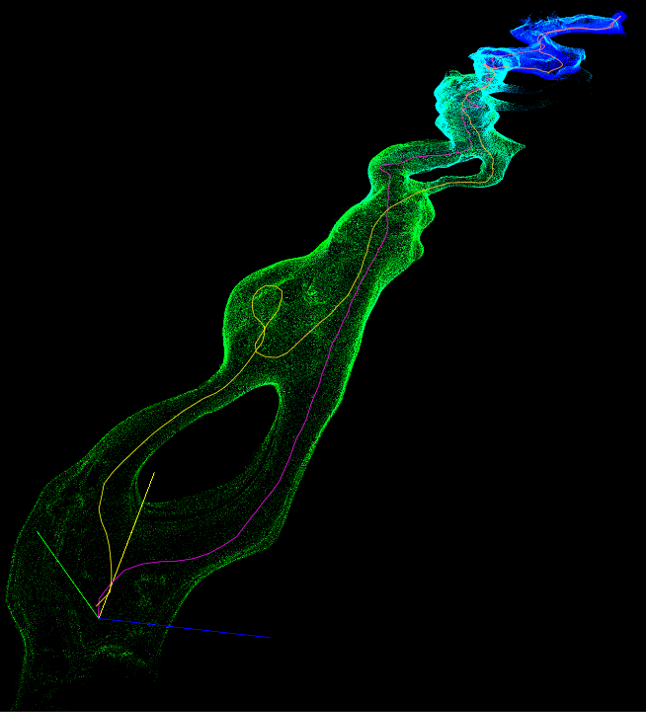}
\end{subfigure}
\caption{Extreme environments successfully explored and mapped using various robotic platforms.  From left to right:  Husky in a coal mine environment, Spot in a power plant environment, Spot in a lava-tube cave environment.  Colors of points (blue to green) indicate z height.  Purple to yellow line indicates path traveled by robot.  Blue/Yellow/Green axes indicate 10m scale.  }
\label{fig:maps}
\vspace{-5mm}
\end{figure}

Figure \ref{fig:maps} shows $\approx 1$ hour of autonomous exploration and mapping in various extreme environments.  These environments include the locales for the DARPA Subterranean Challenge.  Utilizing our resilient navigation methods, both wheeled and legged robots were able to successfully navigate autonomously and in comm-denied environments over kilometer-range distances.


\subsection{Lessons Learned and Future Work}
The following is a summary of the lessons learnt from our intense test campaign; to stem future research:
\begin{itemize}
\item \textbf{Blind Spots in Sensing:}

Our architecture uses instantaneous depth measurements for generating optimal local kinodynamic trajectories in the robot's ego frame. 
While this prevents any localization noise from percolating into the costmap, it requires large number of sensors to ensure sufficient coverage and density.
Violation of this requirement can result in false negative obstacles especially in the blind-spots of the sensors.
This problem can be mitigated with better sensor coverage or perception-aware planning.

\item \textbf{Very Small Obstacles and Narrow Passages:} 

It is challenging to detect obstacles whose size is comparable to the noise levels of the depth sensors. 
Similar challenge exists while navigating through passages whose size is very close to the robot width.
This issues can be potentially resolved with better ground clearance and smaller width of the robot.

\item \textbf{Non-Geometric Hazards:}
There were a few cases where a robot was immobilized due to non-geometric hazards such as mud and water. These hazards mainly triggered slips on the ground contact points and made the wheeled robots get stuck and the legged robots fall. 
One potential solution to this problem is the use of machine learning techniques to detect these non-geometric hazards.

\end{itemize}

\section{Conclusion}
\label{sec:conclusion}

A resilient solution to navigation in environments with extreme traversability and perceptual conditions (Fig.~\ref{fig:traversability_challenges}) was proposed by leveraging a multi-fidelity traversability estimation using a combination of instantaneous depth scans and occupancy belief maps constructed using localization that exploited redundancy and heterogeneity in sensing.
The proposed solution was adaptive, real-time, on-board and works without any prior maps or GPS.
Single-point failures were further eliminated by using health monitoring and recovery behaviors.
Through intense field test campaigns, we demonstrated that the proposed approach scales to different environments and four different types of robots.
Future work includes addressing non-geometric hazards such as mud, dirt, etc. and testing on platforms with higher speed.

\begin{acknowledgement}
The authors would like to thank Anushri Dixit and Joel Burdick for support with the Ackermann robot and members of Team CoSTAR for their hardware and testing support.
The work is partially supported by the Jet Propulsion
Laboratory, California Institute of Technology, under a contract with the National Aeronautics and Space Administration (80NM0018D0004), and Defense Advanced Research
Projects Agency (DARPA).
\end{acknowledgement}

\bibliographystyle{ieeetr}
\bibliography{tunnel_traversability.bib}
\end{document}